\pdfoutput=1

\documentclass[11pt]{article}

\usepackage{latex/acl}

\usepackage{times}
\usepackage{latexsym}
\usepackage{hyperref}
\usepackage{url}
\usepackage{bm}
\usepackage{algorithmic}
\usepackage{algorithm}
\usepackage{graphicx}
\usepackage{tabularx}
\usepackage{amsmath}
\usepackage{booktabs}
\usepackage{xcolor}
\usepackage{stfloats}
\usepackage{hwemoji}

\usepackage[T1]{fontenc}

\usepackage[utf8]{inputenc}

\usepackage{microtype}

\usepackage{inconsolata}

\newcolumntype{C}{>{\centering\arraybackslash}X}
\newcolumntype{L}{>{\raggedright\arraybackslash}X}
\newcolumntype{R}{>{\raggedleft\arraybackslash}X}

\title{Unlearning Traces the Influential Training Data of Language Models}

\author{
Masaru Isonuma~\textsuperscript{1,2}
\quad
Ivan Titov~\textsuperscript{1,3} \\
\textsuperscript{1}University of Edinburgh
\quad
\textsuperscript{2}University of Tokyo 
\quad
\textsuperscript{3}University of Amsterdam\\
\texttt{m.isonuma@ed.ac.uk}
\quad
\texttt{ititov@inf.ed.ac.uk}
}

\begin{document}
\maketitle
\begin{abstract}
Identifying the training datasets that influence a language model's outputs is essential for minimizing the generation of harmful content and enhancing its performance.  
Ideally, we can measure the influence of each dataset by removing it from training; however, it is prohibitively expensive to retrain a model multiple times.
This paper presents UnTrac: \emph{Un}learning \emph{Trac}es the influence of a training dataset on the model's performance.
UnTrac is extremely simple; each training dataset is unlearned by gradient ascent, and we evaluate how much the model's predictions change after unlearning.
Furthermore, we propose a more scalable approach, UnTrac-Inv, which unlearns a test dataset and evaluates the unlearned model on training datasets.
UnTrac-Inv resembles UnTrac, while being efficient for massive training datasets.
In the experiments, we examine if our methods can assess the influence of pretraining datasets on generating toxic, biased, and untruthful content.
Our methods estimate their influence much more accurately than existing methods while requiring neither excessive memory space nor multiple checkpoints.
\end{abstract}

\section{Introduction}

\begin{figure*}[t!]
\centering
\includegraphics[width=\linewidth]{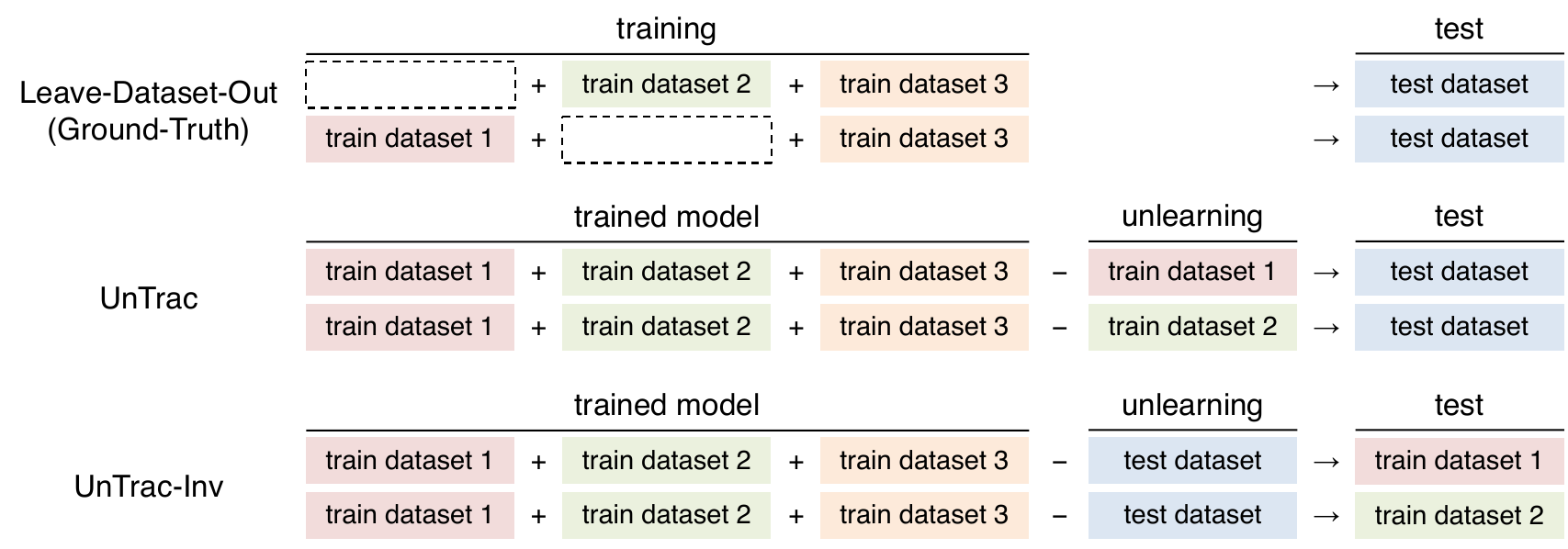}
\caption{Overview of leave-dataset-out vs. proposed methods, UnTrac and UnTrac-Inv.}
\label{fig:introduction}
\end{figure*}

Large language models (LLMs) have had a significant impact on our society. 
They exhibit remarkable abilities (e.g., chain-of-thought reasoning) without being explicitly trained for such tasks.
At the same time, LLMs also pose potential risks, such as the amplification of discrimination through the propagation of toxic language.
LLMs are trained on a vast number of corpora via pretraining or refined through finetuning on diverse tasks.
Although some efforts have been made to unravel the black box of LLMs \cite[e.g.,][]{grosse2023studying, feng-etal-2023-pretraining}, it is still unclear which data sources cause their unprecedented abilities and potential harms.

Ideally, we can answer this question by removing each dataset from the training datasets and assessing the change in the model's performance after retraining (leave-dataset-out).
However, since we need to retrain a model on each dataset, leave-dataset-out is prohibitively expensive.
Training data attribution overcomes this problem by approximating the influence with Hessian-based influence functions \cite[HIF;][]{koh2017understanding, koh2019accuracy} or tracking changes in test loss during training \cite[TracIn;][]{pruthi2020estimating}.
However, HIF requires a large memory space to approximate the inverse Hessian \cite{schioppa2022scaling}, while TracIn generally needs multiple model checkpoints.

In this paper, we propose UnTrac, which traces the influence of a training dataset by unlearning it from a trained model (Figure \ref{fig:introduction}).
Leave-dataset-out removes each training dataset and measures its influence by assessing the trained model's performance on a test dataset.
Analogous to leave-dataset-out, UnTrac unlearns each training dataset and estimates its influence by assessing the unlearned model's performance on a test dataset.
Unlearning has been studied to eliminate sensitive data from a trained model \cite{cao2015towards, ginart2019making, mehta2022deep} and has recently been applied to LLMs \cite{jang-etal-2023-knowledge, chen-yang-2023-unlearn}.
Following \citet{jang-etal-2023-knowledge}, we unlearn a training dataset using gradient ascent, in contrast to the gradient descent normally used in training. Interestingly, \citet{schioppa2023theoretical} argued that influence functions can be regarded as an approximation of the effect of finetuning on a number of examples (e.g., unlearning mislabeled examples). With UnTrac, instead of using the approximations, we directly quantify the effect of unlearning.  

When many datasets are used for training, UnTrac is computationally costly because unlearning must be run for every individual training dataset.
To overcome this drawback, we propose UnTrac-Inv as a scalable approach particularly effective for an increasing number of training datasets.
UnTrac-Inv ``unlearns'' a test dataset instead of training datasets and evaluates the unlearned model on training datasets.
UnTrac-Inv requires only a single run of unlearning, and, as we will show, can be considered as an efficient approximation of UnTrac.

In our experiments, we first examine whether our methods can trace influential training tasks in the setting of finetuning. 
We created a dataset representing a mixture of synthetic tasks, designed to evaluate our method's capability in assessing the influence of each task.
In order to make this assessment more challenging, we have created task pairs that, while semantically distinct, require responses in the same format.
Additionally, we include pairs that are nearly identical in content but demand responses in differing formats. 
By estimating the influence across these task pairs, we verify that our methods are not overly reliant on superficial similarities between tasks. 
In this controlled dataset, we show that our methods accurately assess the influence of training tasks, where we use the expensive leave-dataset-out method as the ground-truth, and are only slightly affected by the output format.

Next, we assess whether our methods can identify the source of harmful content generated by a pretrained language model.
Using smaller open pretrained transformers \cite[OPT-125M;][]{zhang2022opt}, the influence of eight pretraining datasets is estimated.
We use three test datasets: Toxigen \cite{hartvigsen-etal-2022-toxigen}, WinoBias \cite{zhao-etal-2018-gender}, and TruthfulQA \cite{lin-etal-2022-truthfulqa}, which contain toxic language, biased text, and false answers to various questions, respectively.
We calculate the ground-truth influence of each training dataset and evaluate the correlation between the estimated influence and ground-truth influence.
We demonstrate that our methods accurately estimate the influence of pretraining datasets, significantly outperforming other influence functions.

Finally, we investigate how hyperparameters affect the performance of our methods.
We found that UnTrac works robustly as long as we use preconditioned gradient methods with higher learning rates and a sufficient number of training iterations.
In contrast, UnTrac-Inv works well for large batch sizes while being relatively sensitive to the learning rate and the number of training steps.

\section{Problem Formulation}
\label{sec:formulation}

Our goal is to estimate the influence of a training dataset on the model's predictions on a test dataset.
To formalize this goal, we assume the counterfactual that a model is trained on the mixture of all datasets except for a dataset $\mathcal{Z}$: $\bm{\theta}_{-\mathcal{Z}}$. 
The ground-truth influence of the training dataset $\mathcal{Z}$ on a test dataset $\mathcal{Z}'$ is defined as Equation \eqref{eq:ground_truth} using model $\bm{\theta}_{-\mathcal{Z}}$ and $\bm{\theta}_0$, which is trained on all datasets $\mathcal{D}$.
\begin{equation}
\label{eq:ground_truth}
\begin{aligned}[c]
\!\mathcal{I}_{\mathrm{truth}}(\mathcal{Z}', \mathcal{Z}) 
&\!=\! \sum_{j=1}^{N'} L(z'_{j}, \bm{\theta}_{-\mathcal{Z}}) \!-\! L(z'_{j}, \bm{\theta}_{0})
\end{aligned}
\end{equation}
where $z'_{j}$ is the $j$-th batch in the test dataset $\mathcal{Z}'$, $N'$ is the number of test batches, and $L$ is the loss function.
\citet{koh2019accuracy} use all examples in the dataset to train the counterfactual model: $\bm{\theta}_{-\mathcal{Z}} \!=\! \mathrm{arg} \min_{\bm{\theta}} \sum_{z \in \mathcal{D} \setminus \mathcal{Z}} L(z, \bm{\theta})$.
This definition overly emphasizes the influence of large datasets, as, when removing them, the number of training examples drops substantially.
However, when asking about the influence of a dataset, our primary interest often centers on whether the type of data present within a dataset wields considerable influence.
We modify the definition of dataset's influence so as to better align with this research question.
Thus, for every training dataset $\mathcal{Z}$, we train a model for the same number of training steps $T$ instead of the entire dataset: $\bm{\theta}_{-\mathcal{Z}} = \mathrm{arg} \min_{\bm{\theta}} \sum_{t=1}^{T} L(z_t, \bm{\theta})$ where $z_t \sim \mathcal{D} \setminus \mathcal{Z}$.
This setup is practical for evaluating the influence of datasets of different sizes.\footnote{In Appendix \ref{apx:loo}, we show the conventional leave-dataset-out setup overestimates the influence of large datasets.}

\section{Methods}

\subsection{UnTrac}
\label{sec:untrac}

Here, we formally introduce UnTrac, which estimates the influence of a training dataset on a test dataset by unlearning.
Let $\mathcal{Z}'$ be a test dataset, $\bm{\theta}_{i}$ be the model parameters after the $i$-th unlearning step ($\bm{\theta}_{0}$ is the trained model parameters).
The influence of a training dataset $\mathcal{Z}$ on a test dataset $\mathcal{Z}'$ is defined as the change in test loss due to unlearning.
\begin{equation}
\label{eq:influence}
\begin{aligned}
\mathcal{I}(\mathcal{Z}', \mathcal{Z}) 
&\!=\! \sum_{j=1}^{N'} L(z'_{j}, \bm{\theta}_{T}) \!-\! L(z'_{j}, \bm{\theta}_{0}) \\
&\!=\! \sum_{j=1}^{N'} \sum_{i=1}^{T} L(z'_{j}, \bm{\theta}_{i}) \!-\! L(z'_{j}, \bm{\theta}_{i-1})
\end{aligned}
\end{equation}
where $T$ is the number of unlearning steps.
Here, $\bm{\theta}_{i}$ is updated via gradient ascent, which \emph{maximizes} the loss of the $i$-th batch $z_{i}$ in the training dataset $\mathcal{Z}$.
If we use stochastic gradient ascent, the updated parameters can be written as Equation \eqref{eq:gradient_ascent}; however, any optimizer can be used for unlearning.
\begin{equation}
\label{eq:gradient_ascent}
\begin{aligned}
\bm{\theta}_{i} = \bm{\theta}_{i-1} + \eta_i \nabla_{\bm{\theta}}L(z_{i}, \bm{\theta}_{i-1})
\end{aligned}
\end{equation}

\subsection{UnTrac-Inv}
\label{sec:untrac_reverse}

When many datasets are used for training, UnTrac is computationally costly because unlearning must be run for every training dataset.
In a practical scenario, we are interested in detecting which training dataset influences a particular test dataset.
Here, we introduce UnTrac-Inv, an alternative scalable approach that can handle an increasing number of training datasets.
UnTrac-Inv unlearns the test dataset instead of training datasets and measures the change in loss on the training datasets.
UnTrac-Inv computes the influence by Equation \eqref{eq:influence_inv}.
\begin{equation}
\begin{aligned}
\label{eq:influence_inv}
\mathcal{I'}(\mathcal{Z}', \mathcal{Z}) 
&\!= \sum_{i=1}^{N} L(z_{i}, \bm{\theta}_{T'}) - L(z_{i}, \bm{\theta}_{0}) \\
&\!= \sum_{i=1}^{N} \sum_{j=1}^{T'} L(z_{i}, \bm{\theta}_{j}) - L(z_{i}, \bm{\theta}_{j-1})
\end{aligned}
\end{equation}
where $N$ is the number of batches in the training dataset, and $T'$ is the number of unlearning steps.

This alternative influence can be regarded as an approximation to UnTrac, Equation (\ref{eq:influence}).
Note that $\bm{\theta}_{j} = \bm{\theta}_{j-1} + \eta_j \nabla_{\bm{\theta}}L(z'_{j}, \bm{\theta}_{j-1})$, and we use the first-order approximation $L(z_{i}, \bm{\theta}_{j}) - L(z_{i}, \bm{\theta}_{j-1}) \approx \nabla_{\bm{\theta}}L(z_{i}, \bm{\theta}_{j-1}) (\bm{\theta}_{j} - \bm{\theta}_{j-1})$. 
Equation \eqref{eq:influence} and \eqref{eq:influence_inv} can then be re-approximated as Equation \eqref{eq:approx_influence} and \eqref{eq:approx_influence_inv}, respectively.
\begin{equation}
\label{eq:approx_influence}
\begin{aligned}
&\mathcal{I}(\mathcal{Z}', \mathcal{Z}) \\
&\!\approx \sum_{i=1}^{T} \sum_{j=1}^{N'} \eta_i \nabla_{\bm{\theta}}L(z_{i}, \bm{\theta}_{i-1})^{\top} \nabla_{\bm{\theta}}L(z'_{j}, \bm{\theta}_{i-1}) \\
\end{aligned}
\end{equation}
\begin{equation}
\label{eq:approx_influence_inv}
\begin{aligned}
&\mathcal{I'}(\mathcal{Z}', \mathcal{Z}) \\
&\!\approx \sum_{i=1}^{N} \sum_{j=1}^{T'} \eta_j \nabla_{\bm{\theta}}L(z_{i}, \bm{\theta}_{j-1})^{\top} \nabla_{\bm{\theta}}L(z'_{j}, \bm{\theta}_{j-1})
\end{aligned}
\end{equation}
If the number of unlearning steps is one $(T\!=T'\!=1)$, and a single batch contains all examples, $\mathcal{I}(\mathcal{Z}', \mathcal{Z})$ corresponds to $\mathcal{I'}(\mathcal{Z}', \mathcal{Z})$.
This suggests that UnTrac-Inv should work well with a small number of unlearning steps and a large batch size.
We will empirically validate it later in Section \ref{sec:epoch}.

\section{Relation to Other Influence Functions}
\subsection{TracIn, GradDot \& GradCos}
\label{sec:tracin}

\citet{pruthi2020estimating} proposed TracIn, which traces the influence of a training example by the total change in test loss during training.
While the original TracIn assesses the influence of individual training examples, it can be easily extended to assess a whole dataset.
As accounting for the loss reduction at every step is computationally expensive, they approximate it using model checkpoints, assuming that every example in the training dataset is encountered once between these checkpoints.
By approximating the loss reduction with gradients at each checkpoint $t$: $L(z'_{j}, \bm{\theta}_{t}) - L(z'_{j}, \bm{\theta}_{t+1}) \approx \eta_t \nabla_{\bm{\theta}}L(z_{i}, \bm{\theta}_{t})^{\top} \nabla_{\bm{\theta}}L(z'_{j}, \bm{\theta}_{t})$, TracIn is defined as:
\begin{equation}
\begin{aligned}
\label{eq:tracin_cp}
&\text{TracIn}(\mathcal{Z}', \mathcal{Z}) \\
&\!= \sum_{t \in \mathcal{T}_{cp}} \sum_{i=1}^{N} \sum_{j=1}^{N'} \eta_t \nabla_{\bm{\theta}}L(z_{i}, \bm{\theta}_{t})^{\top} \nabla_{\bm{\theta}}L(z'_{j}, \bm{\theta}_{t}) \\
\end{aligned}
\end{equation}
where $\mathcal{T}_{cp}$ denotes the set of training steps where the model checkpoints are saved.
As using multiple checkpoints induces substantial overhead, only the last checkpoint is often used in practice \cite{schioppa2023theoretical}, which is referred to as GradDot.
\citet{barshan2020relatif} pointed out that some outlier training examples have significantly large gradients, leading to an overestimation of their influences.
Therefore, normalizing the gradients (i.e., replacing the dot product with cosine similarity) can be effective, referred to as GradCos \cite{han-tsvetkov-2021-influence-tuning, akyurek-etal-2022-towards}.

\subsection{Hessian-based Influence Functions} 
\label{sec:influence}

Hessian-based influence functions (HIF) are grounded in robust statistics \cite{hampel1974influence, cook1982residuals} and were introduced to deep learning by \citet{koh2017understanding}.
\citet{koh2019accuracy} used HIF to assess the influence of multiple training examples, and the estimated influence correlates well with the ground-truth.
Given a trained model $\bm{\theta}_0$, HIF estimates the influence of a training dataset $\mathcal{Z}$ on a test dataset $\mathcal{Z}'$ as Equation \eqref{eq:influence_functions}.
\begin{equation}
\begin{aligned}
\label{eq:influence_functions}
&\text{HIF}(\mathcal{Z}', \mathcal{Z}) \\
&= \sum_{i=1}^{N} \sum_{j=1}^{N'} \nabla_{\bm{\theta}}L(z_{i}, \bm{\theta}_0)^{\top} \bm{H}_{\bm{\theta}}^{-1} \nabla_{\bm{\theta}}L(z'_{j}, \bm{\theta}_0)
\end{aligned}
\end{equation}
where $\bm{H}_{\bm{\theta}}$ is the Hessian of training loss: $\bm{H}_{\bm{\theta}} = 1/N \sum_{i=1}^{N} \nabla_{\bm{\theta}}^2 L(z_{i}, \bm{\theta}_0)$.

\subsection{Connection to UnTrac \& UnTrac-Inv}
\label{sec:connection}

GradDot, GradCos, and HIF can be viewed as special cases of UnTrac and UnTrac-Inv.
Suppose a single training batch contains all training examples.
The increase in test loss after a single step of unlearning can be approximated as Equation \eqref{eq:unlearn_single}.
\begin{equation}
\begin{aligned}
\label{eq:unlearn_single}
\!\mathcal{I}(\mathcal{Z}', \mathcal{Z})
&\!=\! \sum_{j=1}^{N'} L(z'_{j}, \bm{\theta}_1) - L(z'_{j}, \bm{\theta}_0) \\
&\approx \sum_{j=1}^{N'} \nabla_{\bm{\theta}}L(z'_{j}, \bm{\theta}_0)^{\top} (\bm{\theta}_1 - \bm{\theta}_0)
\end{aligned}
\end{equation}
When we use SGD, Equation \eqref{eq:unlearn_single} corresponds to GradDot by substituting $\bm{\theta}_1 - \bm{\theta}_0 = \eta_0 \sum_{i=1}^{N} \nabla_{\bm{\theta}}L(z_{i}, \bm{\theta}_{0})$.
Similarly, It corresponds to GradCos if we use RMSProp or Adam: $\bm{\theta}_1 - \bm{\theta}_0 = \eta_0 \sum_{i=1}^{N} \nabla_{\bm{\theta}}L(z_{i}, \bm{\theta}_{0}) / \| \sum_{i=1}^{N} \nabla_{\bm{\theta}}L(z_{i}, \bm{\theta}_{0}) \|$, and corresponds to HIF if we use Newton's method: $\bm{\theta}_1 - \bm{\theta}_0 = \sum_{i=1}^{N} \bm{H}_{\bm{\theta}}^{-1} \nabla_{\bm{\theta}}L(z_{i}, \bm{\theta}_{0})$.
In the same way, these influence functions can also be regarded as a special case of UnTrac-Inv.
HIF can be regarded as providing an approximation to UnTrac.

\section{Experiments}
\label{sec:experiment}

We evaluate if our methods can measure the influence of training datasets across different model architectures (encoder-decoder and decoder-only) and training setups (pretraining and fine-tuning).\footnote{The code and data are available at: \url{https://github.com/misonuma/untrac/}. More implementation details are noted in Appendix \ref{sec:implementation}.}

Since we would not generally have a validation set (i.e., ground-truth attributions), it is impractical to tune the hyperparameters for each experiment. 
Thus, we set the hyperparameters based on the experimental results on Toxigen, one of the test datasets used in Section \ref{sec:pretraining}.
The same hyperparameters are used across all the experiments.
We use Adam \cite{kingma2014adam} with a constant learning rate of 5e-5, $\beta_1$ = 0.9, and $\beta_2$ = 0.999, while Adafactor is used in Section \ref{sec:instructiontuning} due to memory constraints.
Gradient clipping is turned off during unlearning.
As for UnTrac, we set the batch size to 1 and run unlearning for 1 epoch.
As discussed in Section \ref{sec:untrac_reverse}, UnTrac-Inv requires a small number of unlearning steps and a large batch size.
Hence, we set the batch size as 256 (a single batch contains all test examples) and the number of unlearning steps (epochs) as 5.
In Section \ref{sec:discussion}, we discuss the hyperparameter sensitivity of our methods.

The baseline methods are as follows:

\paragraph{TracIn, GradDot \& GradCos} 

We compute the gradient w.r.t. all of the model parameters.
TracIn uses the checkpoints saved for every 128 steps in Section \ref{sec:instructiontuning} and 10,000 steps in Section \ref{sec:pretraining}. 

\paragraph{HIF}
Following \citet{koh2017understanding}, we approximate the inverse Hessian by LISSA \cite{agarwal2017second}, where the number of iterations is set to 10.
We also use Arnoldi iteration with low-rank eigenvector projection following \citet{schioppa2022scaling}.
We set the number of Arnoldi iterations to $n$ = 25 and the number of eigenvectors to $\tilde{p}$ = 25 in Section \ref{sec:instructiontuning} due to the memory constraints, while $n$ = 200 and $\tilde{p}$ = 100 in Section \ref{sec:pretraining}.
Following the previous studies, the gradients of training datasets are normalized, and 256 training examples are used to approximate the Hessian of training loss.

\begin{table*}[t!]
\centering
\small
\begin{tabularx}{\textwidth}{p{3.3cm}p{10.5cm}L} 
\toprule
Synthetic Datasets A&Input&Output\\ 
\midrule
Test (Task P/Format P)&What is the number that comes after \{0\}?&\{1\} \\
\midrule
Train1 (Task P/Format P)&Determine the number that succeeds \{two\}. Provide your answer in numerical form.&\{3\} \\
Train2 (Task P/Format Q)&Determine the number that succeeds \{one\}. Provide your answer in words.&\{two\} \\
Train3 (Task Q/Format P)&Determine the length of `\{problem\}'. Provide your answer in numerical form.&\{7\} \\
Train4 (Task Q/Format Q)&Determine the length of `\{align\}'. Provide your answer in words.&\{five\} \\
\bottomrule
\end{tabularx}
\end{table*}
\begin{table*}[t!]
\small
\begin{tabularx}{\textwidth}{p{3.3cm}p{10.5cm}L}
\toprule
Synthetic Datasets B&Input&Output\\ 
\midrule
Test (Task P/Format P)&What letter remains in `\{xmais\}' after extracting `\{s\}', `\{x\}', `\{m\}', `\{i\}'?&\{a\} \\
\midrule
Train1 (Task P/Format P)&Identify the character left after removing `\{0\}', `\{1\}', `\{4\}', `\{7\}' from `\{71b40\}'.&\{b\} \\
Train2 (Task P/Format Q)&Identify the character left after removing `\{6\}', `\{7\}', `\{5\}', `\{2\}' from `\{27516\}'.&\{1\} \\
Train3 (Task Q/Format P)&Identify the part of speech of `\{problem\}'. Select your answer with the associated letter.   Choices:   a. noun  b. verb  c. adjective  d. adverb&\{a\} \\
Train4 (Task Q/Format Q)&Identify the part of speech of `\{align\}'. Select your answer with the associated number.   Choices:   0. noun  1. verb  2. adjective  3. adverb&\{1\} \\
\bottomrule
\end{tabularx}
\caption{Example of the synthetic datasets A (top) and B (bottom). \{Strings in braces\} are varied with each example.}
\label{tbl:synthetic_dataset}
\end{table*}

\begin{table*}[t!]
\centering
\small
\begin{tabularx}{\textwidth}{p{1.8cm}RRRRp{0.01cm}RRRR}
\toprule
&\multicolumn{4}{c}{Synthetic Datasets A}& &\multicolumn{4}{c}{Synthetic Datasets B}\\ 
\cmidrule{2-5} \cmidrule{7-10}
Train Dataset & 1{\scriptsize (TaskP/FmtP)} & 2{\scriptsize (TaskP/FmtQ)} & 3{\scriptsize (TaskQ/FmtP)} & 4{\scriptsize (TaskQ/FmtQ)} & & 1{\scriptsize (TaskP/FmtP)} & 2{\scriptsize (TaskP/FmtQ)} & 3{\scriptsize (TaskQ/FmtP)} & 4{\scriptsize (TaskQ/FmtQ)} \\
\midrule
GradDot & 0.497 & 0.930 & -1.680 & 0.253 &  & -0.561 & 1.730 & -0.655 & -0.513 \\
GradCos & 0.993 & 0.573 & -1.637 & 0.071 &  & 1.197 & 0.708 & -1.307 & -0.598 \\
HIF (Arnoldi) & 1.059 & 0.566 & -0.026 & -1.599 &  & 1.368 & 0.015 & 0.074 & -1.457 \\
HIF (LISSA) & 0.156 & -1.674 & 0.658 & 0.860 &  & -0.600 & 1.708 & -0.785 & -0.323 \\
TracIn & 0.090 & 1.575 & -1.083 & -0.582 &  & -0.174 & 1.638 & -1.060 & -0.404 \\
\emph{UnTrac} & 1.443 & 0.344 & -0.574 & -1.212 &  & 1.492 & 0.175 & -0.414 & -1.253 \\
\emph{UnTrac-Inv} & 0.980 & 1.014 & -1.104 & -0.891 &  & 1.056 & 0.905 & -0.716 & -1.245 \\
\midrule
Ground Truth & 1.263 & 0.692 & -1.064 & -0.891 &  & 1.150 & 0.837 & -1.025 & -0.962 \\
\bottomrule
\end{tabularx}
\caption{Influence of the training datasets estimated by each method. The average is shown across four runs. The values are standardized for each method to normalize the range of values. ``Fmt'' is the abbreviation of ``Format''.}
\label{tbl:synthetic_result}
\end{table*}

\subsection{Tracing Influential Training Tasks}
\label{sec:instructiontuning}
We examine whether UnTrac can detect influential training tasks in the setting of finetuning.
If a task is unlearned from a model, one may hypothesize that the model will no longer respond in the format required by the task, regardless of the input.
Hence, we are concerned that unlearning not relevant tasks but those having the same answer format as a test task may adversely affect the test task's performance.
This may not be consistent with the leave-dataset-out ground-truth and, thus, lead to an overestimation of the non-relevant task's influence.
Using synthetic datasets, we show that UnTrac can assess the influence of training tasks properly, regardless of the output format.

\paragraph{Model}
We use T5 \cite{raffel2020exploring} as a pretrained encoder-decoder model.
Specifically, we use the LM-adapted T5-XL (3B), which is finetuned on language modeling \cite{lester-etal-2021-power}.\footnote{\url{https://huggingface.co/google/t5-xl-lm-adapt}}

\paragraph{Dataset}

We create two synthetic datasets, each containing one test dataset and four training datasets.
The test datasets contain examples of task P with the output format P (Task P/Format P).
The task and output format of training datasets are set with respect to the test dataset as follows:
\begin{enumerate}
\setlength{\parskip}{0pt}
\setlength{\itemsep}{0pt}
\item Training dataset 1 (Task P/Format P): similar task P with the same output format P.
\item Training dataset 2 (Task P/Format Q): similar task P with different output format Q.
\item Training dataset 3 (Task Q/Format P): irrelevant task Q with the same output format P.
\item Training dataset 4 (Task Q/Format Q): irrelevant task Q with different output format Q.
\end{enumerate}

Table \ref{tbl:synthetic_dataset} presents examples of the synthetic datasets A and B.
Each training dataset and test dataset consists of 256 examples.
The model is trained for 512 steps on the mixture of the four training datasets with a batch size of 2.

\begin{figure*}[t!]
\centering
\begin{minipage}{0.495\linewidth}
\centering
\includegraphics[width=\linewidth]{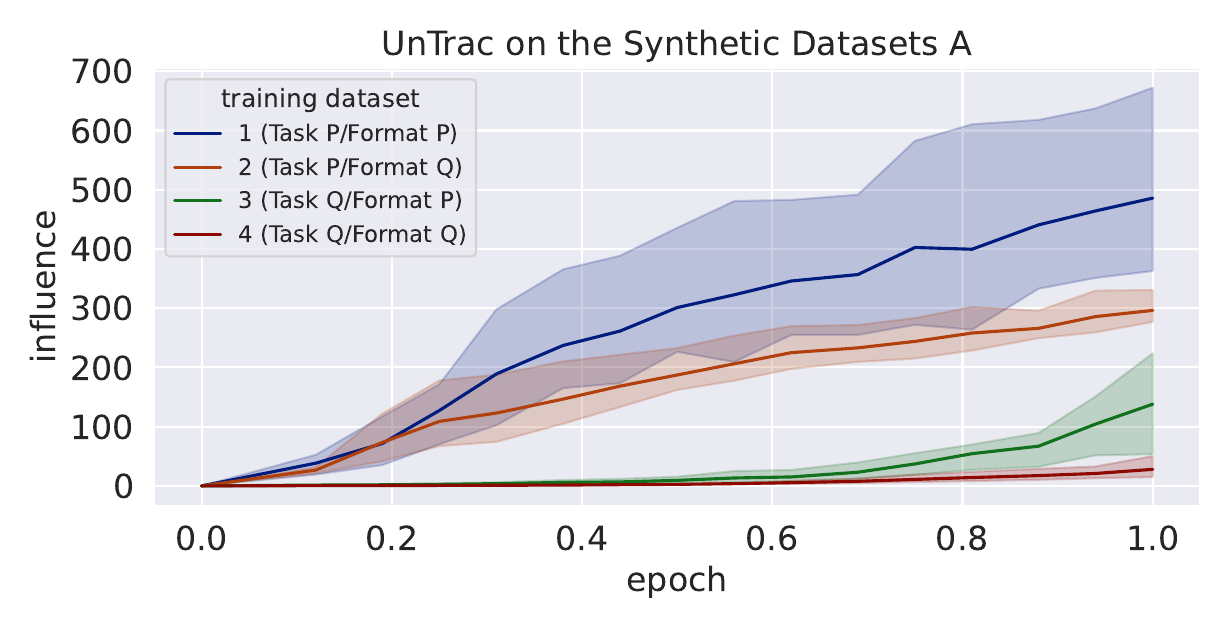}
\end{minipage}
\begin{minipage}{0.495\linewidth}
\centering
\includegraphics[width=\linewidth]{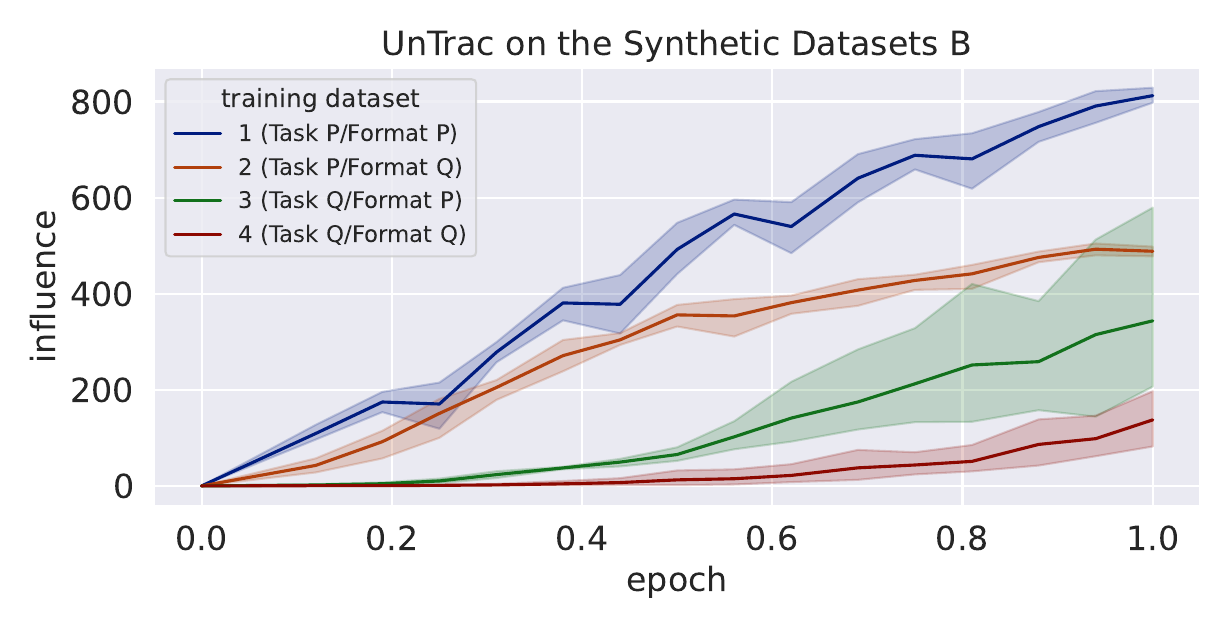}
\end{minipage}
\end{figure*}
\begin{figure*}[t!]
\begin{minipage}{0.495\linewidth}
\centering
\includegraphics[width=\linewidth]{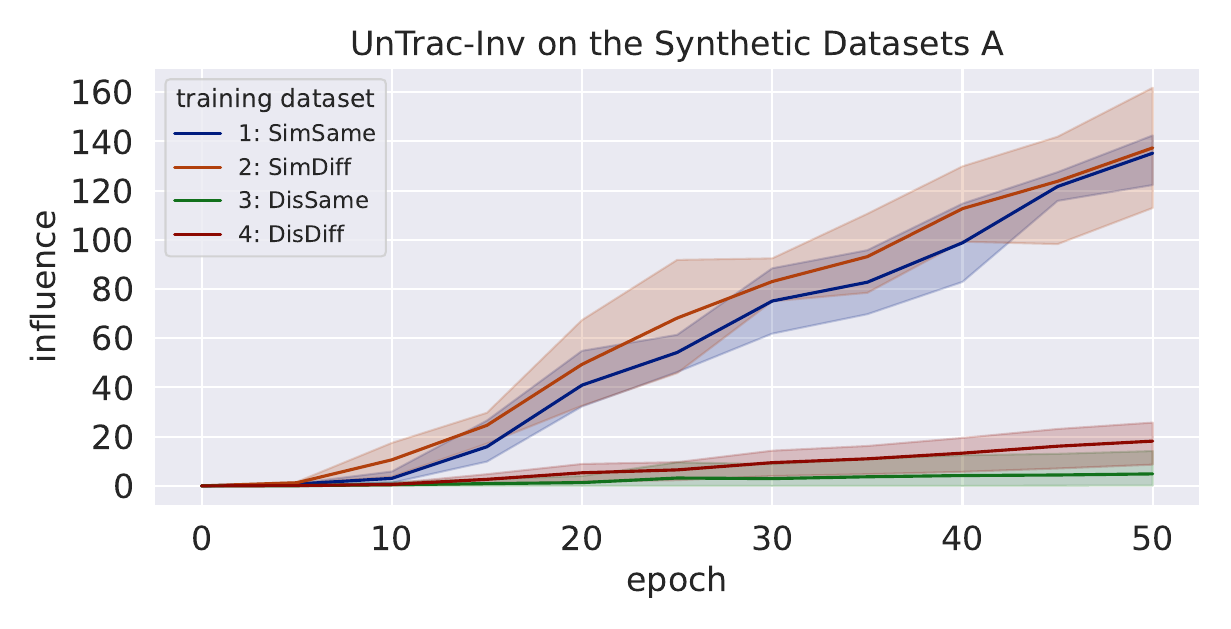}
\end{minipage}
\begin{minipage}{0.495\linewidth}
\centering
\includegraphics[width=\linewidth]{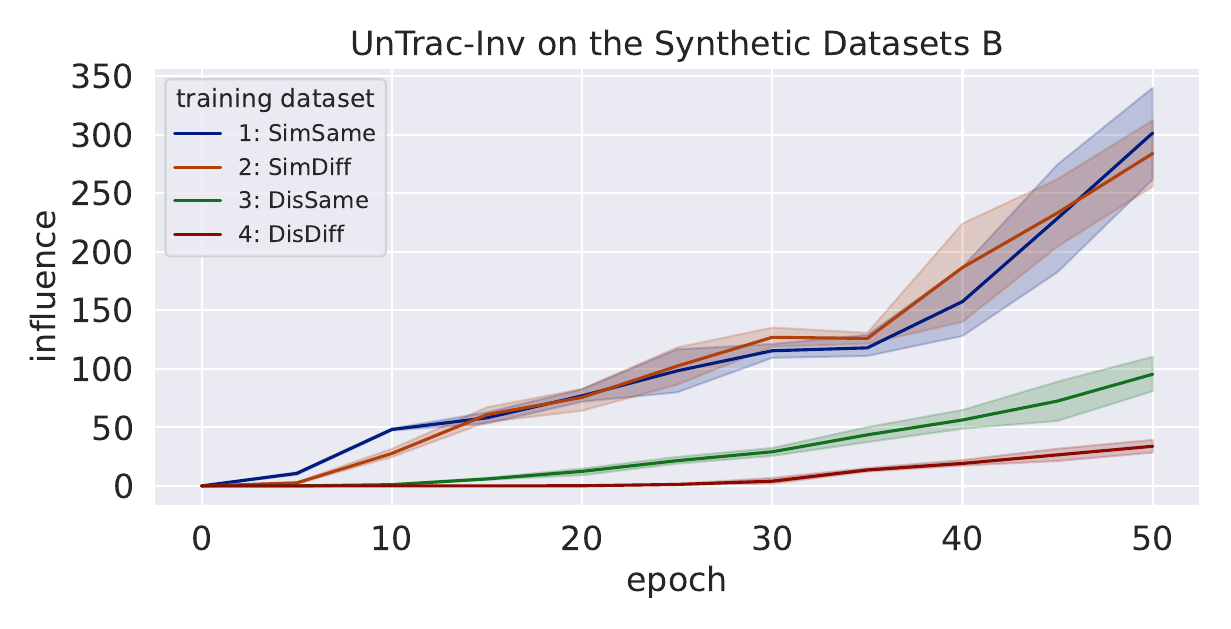}
\end{minipage}
\caption{The influence estimated by UnTrac (top) and UnTrac-Inv (bottom) on the synthetic datasets A (left) and B (right). The line denotes the average across four runs, and the shaded area corresponds to 95\% confidence region.}
\label{fig:unlearn_synthetic}
\end{figure*}

\paragraph{Results}

Table \ref{tbl:synthetic_result} presents the influence of training datasets estimated by each method and the ground-truth influence measured by leave-dataset-out.
The ground truth indicates that datasets 1 and 2 (tasks similar to the test task) are more influential than datasets 3 and 4 (tasks irrelevant to the test task).
The influence estimated by UnTrac and UnTrac-Inv aligns well with the ground-truth influence. 
All other methods assess the influence of dataset 4 as lower than that of other datasets.
However, they tend to overestimate the influence of dataset 3 or underestimate it for datasets 1 and 2.

Figure \ref{fig:unlearn_synthetic} shows the change in influence estimated by UnTrac and UnTrac-Inv on the synthetic datasets A (left) and B (right), respectively.
For both synthetic datasets, our methods estimate the influence of datasets 1 and 2 as greater than that of datasets 3 and 4 across unlearning steps.
These results indicate that UnTrac and UnTrac-Inv appropriately estimate the influence of training datasets in terms of the relevance of tasks and do not seem overly affected by the output format.

\subsection{Tracing Influential Pretraining Corpora}
\label{sec:pretraining}
LLMs sometimes generate toxic, biased, and false content, which must be prevented to safely use language models.
We next verify that our methods can identify the influence of a pretraining dataset on the generation of harmful content.

\paragraph{Model}
We use an open pre-trained transformer language model \cite[OPT;][]{zhang2022opt}.
As computing ground-truth influence for pretraining datasets is expensive, we use a relatively small language model with 125 million parameters.\footnote{\url{https://huggingface.co/facebook/opt-125m}} 
OPT is pretrained for 40,000 steps with a batch size of 8 on the datasets described below.

\paragraph{Dataset}
We use eight pretraining datasets that were used for OPT: BookCorpus \cite{zhu2015aligning}, CC-Stories \cite{trinh2018simple}, CCNewsV2 \cite{liu2019roberta}, and five subsets in the Pile dataset \cite{gao2020pile}: PJ Gutenberg, HackerNews, OpenWebText2, Pile-CC, and Wikipedia.
To investigate whether our methods are effective regardless of the dataset's proportion, we set up two settings: one where each pretraining dataset contains an equal number of examples (40,000) and another where they contain different numbers of examples (Pile-CC: 96,000, OpenWebText2: 64,000, CCNewsV2: 48,000, BookCorpus: 32,000, Stories: 32,000, PJ Gutenberg: 16,000, HackerNews: 16,000, Wikipedia: 16,000).
Each training example consists of a sequence of 1,024 tokens by grouping several examples into a sequence.

In practice, computing influences using the whole pretraining dataset is quite expensive.
Thus, we randomly sample 10,000 examples from each dataset to estimate its influence.
To ensure that the reported results are invariant to the choice of examples, we report the average and standard deviation across four runs using different examples.

Regarding test datasets, we use three datasets: Toxigen \cite{hartvigsen-etal-2022-toxigen}, WinoBias \cite{zhao-etal-2018-gender}, and TruthfulQA \cite{lin-etal-2022-truthfulqa}.
ToxiGen collects machine-generated toxic language against several minority groups.
Winobias contains sentences with entities corresponding to people referred to by their occupation and their pronoun genders.
TruthfulQA comprises questions across several categories and their corresponding untruthful answers.
We measure the negative log-probability of toxic language, biased text, and false answers to compute the influences.

\begin{table*}[t!]
\small
\centering
\begin{tabularx}{\textwidth}{p{1.8cm}RRRp{0.1cm}RRR}
\toprule
&\multicolumn{3}{c}{Equal Training Dataset Size}& &\multicolumn{3}{c}{Different Training Dataset Size}\\ 
\cmidrule{2-4} \cmidrule{6-8}
Test Dataset & ToxiGen & WinoBias & TruthfulQA & & ToxiGen & WinoBias & TruthfulQA \\
\midrule
GradDot & -0.123±0.008 & 0.418±0.018 & 0.156±0.022 &  & -0.250±0.007 & 0.446±0.015 & -0.524±0.003 \\
GradCos & -0.050±0.008 & 0.524±0.014 & 0.447±0.015 &  & -0.337±0.007 & 0.496±0.012 & -0.401±0.004 \\
HIF (Arnoldi) & -0.068±0.023 & 0.559±0.010 & 0.250±0.024 &  & -0.343±0.005 & \textbf{0.584±0.014} & -0.362±0.006 \\
HIF (LISSA) & -0.040±0.328 & 0.389±0.117 & -0.178±0.173 &  & 0.071±0.091 & -0.092±0.042 & -0.098±0.058 \\
TracIn & 0.207±0.010 & 0.082±0.013 & \textbf{0.591±0.014} &  & -0.187±0.005 & 0.183±0.019 & 0.081±0.010 \\
\midrule
\emph{UnTrac} & \textbf{0.419±0.063} & 0.743±0.086 & 0.314±0.223 &  & \textbf{0.403±0.033} & 0.518±0.122 & 0.246±0.082 \\
\emph{UnTrac-Inv} & \textbf{0.372±0.008} & \textbf{0.813±0.012} & 0.582±0.016 &  & \textbf{0.393±0.037} & 0.275±0.125 & \textbf{0.360±0.017} \\
\bottomrule
\end{tabularx}
\caption{Pearson correlation coefficient between the influence estimated by each method and the ground-truth influence computed by leave-dataset-out.
Each figure denotes the mean and standard deviation across four runs.
For each run, we use different 10,000 examples randomly sampled from the training dataset to compute its influence. The highest score and the scores that lie within its standard deviation are highlighted in bold.}
\label{tbl:toxic_result}
\end{table*}

\begin{figure*}[t!]
\centering
\begin{minipage}{0.495\linewidth}
\centering
\includegraphics[width=\linewidth]{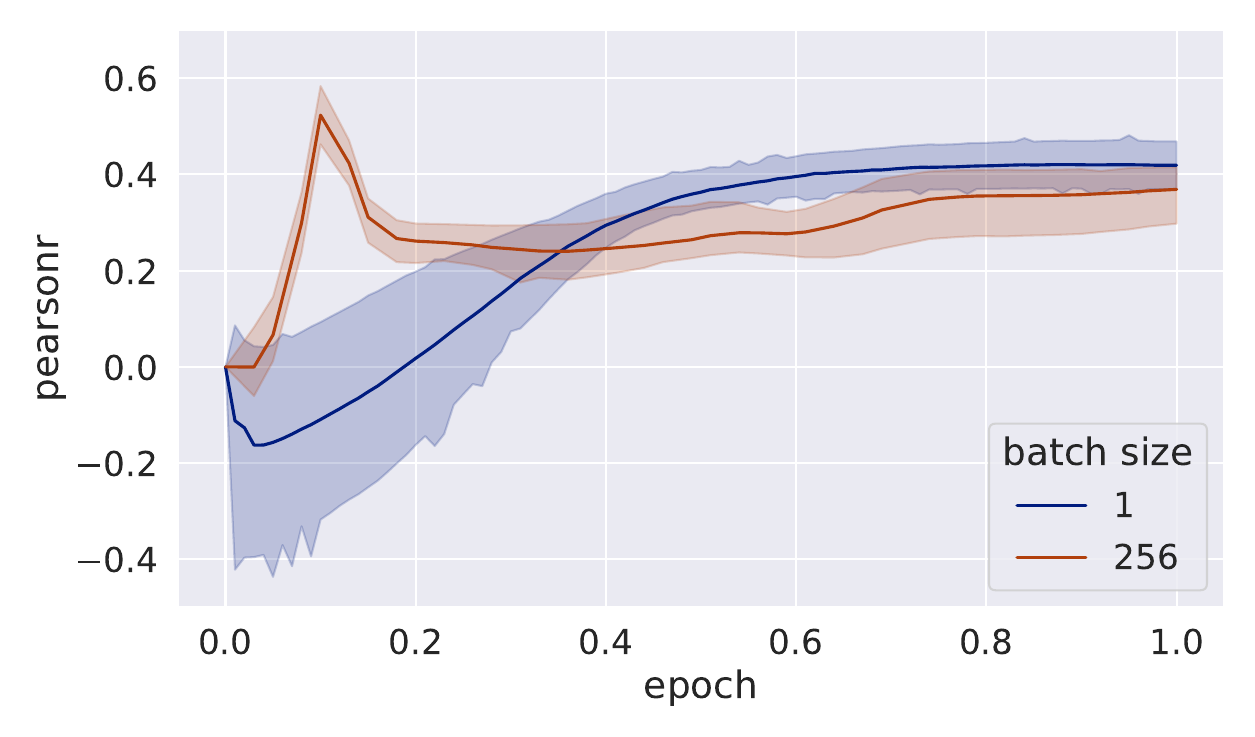}
\end{minipage}
\begin{minipage}{0.495\linewidth}
\centering
\includegraphics[width=\linewidth]{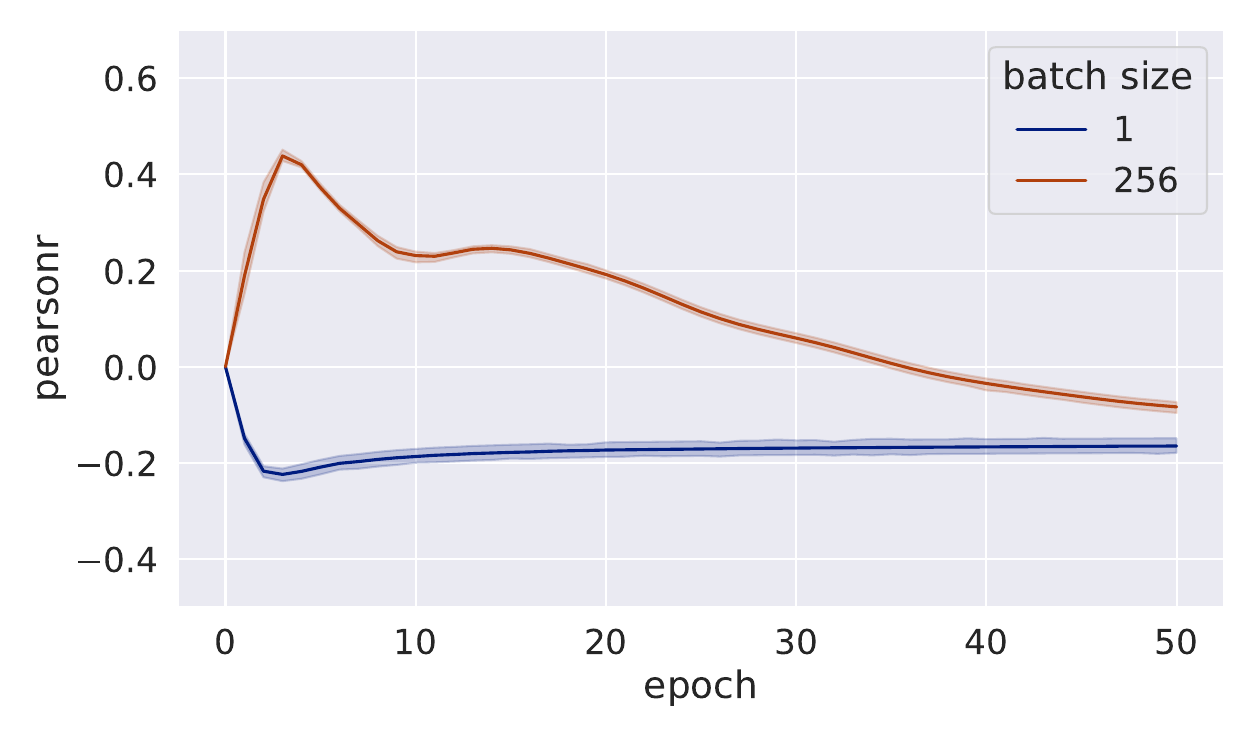}
\end{minipage}
\caption{Pearson correlation coefficient between the ground truth and the influence estimated by UnTrac (left) and UnTrac-Inv (right) over unlearning epochs. 
ToxiGen is used as an evaluation dataset and the size of each training dataset is set to be equal.
The line denotes the average across four runs, and the shaded area corresponds to 95\% confidence region.}
\label{fig:epochs_untrac}
\end{figure*}

\paragraph{Results}

Table \ref{tbl:toxic_result} shows the Pearson correlation coefficient between the estimated influence and the ground truth assessed by leave-dataset-out.\footnote{The same tendency is confirmed when Spearman's rank correlation coefficient is used as a metric (Appendix \ref{apx:spearmanr}).}
As there are only eight pretraining datasets, the correlation between ground truth and estimated influences cannot be statistically significant. 
Therefore, we divided the test dataset into multiple subsets and aggregated the correlation for each subset to enhance the reliability of the results.
We reported the average across 13 subsets for ToxiGen, 4 subsets for WinoBias, and 9 subsets for TruthfulQA.

Across all datasets and settings, the estimated influence by UnTrac and UnTrac-Inv correlates well with the ground truth.
GradCos, GradDot, and HIF (Arnoldi) perform well on Winobias.
However, they show lower performance on Toxigen and TruthfulQA specifically when the dataset sizes are unbalanced.
The performance of HIF (LISSA) is unstable, as indicated by the high variance in its score.
While TracIn achieves a relatively higher correlation with equally sized training datasets, its performance declines when the sizes are different.

We found that the influences estimated by existing methods are not significantly different across training datasets. 
For WinoBias and TruthfulQA, the ground-truth influence of each pre-training dataset does not differ that much. 
Thus, the existing methods achieve high correlations on those datasets as well as ours. 
In contrast, on ToxiGen, the ground-truth influences of pre-training datasets vary significantly; Pile-CC and BookCorpus are more influential than others. 
Our methods can detect these influential datasets, while other approaches fail, leading to higher performance of our methods on ToxiGen.
These results indicate that our methods can robustly trace the influence of pretraining datasets regardless of their proportions.

\section{Discussion}
\label{sec:discussion}

In this section, we explore how hyperparameters affect the performance of our methods.
Following the experiment in Section \ref{sec:pretraining}, we compute the influence of the pretraining dataset on Toxigen under various hyperparameter settings, where the size of each training dataset is set to be equal.
By monitoring the Pearson correlation coefficient between the estimated influence and ground truth, we investigate the hyperparameter sensitivity of our methods.

\subsection{Sensitivity to Epoch \& Batch Size}
\label{sec:epoch}

Figure \ref{fig:epochs_untrac} shows the performance of UnTrac and UnTrac-Inv over unlearning epochs with batch sizes of 1 and 256.
On both batch sizes, UnTrac achieves high and stable performance as the number of unlearning epochs increases.
In contrast, UnTrac-Inv shows entirely different tendencies.
When the batch size is one, UnTrac-Inv performs poorly over the entire run of unlearning. 
When the batch size is 256, the performance of UnTrac-Inv rises for the first several epochs, while it degrades gradually after a while.
As discussed in Section \ref{sec:untrac_reverse}, UnTrac-Inv approximates UnTrac when unlearning is run for a small number of steps with a large batch size.
When the number of unlearning steps is large or the batch size is small, UnTrac-Inv deviates from UnTrac and does not perform well.

These observations may suggest why other influence functions perform worse.
As mentioned in Section \ref{sec:connection}, GradDot, GradCos, and HIF can be regarded as approximations of UnTrac (and UnTrac-Inv) when unlearning is conducted for a single step.
As shown in Figure \ref{fig:epochs_untrac}, UnTrac and UnTrac-Inv underperform when the number of unlearning steps is one.
From the viewpoint of unlearning, a single step is often insufficient to fully trace the influence of training datasets, suggesting why other influence functions underperform. 

\begin{table*}[t!]
\small
\centering
\begin{tabularx}{\linewidth}{p{2cm}RRRRR}
\toprule
Optimizer & SGD & SGD {\scriptsize w/momentum} & RMSProp & Adam & Adafactor \\
\midrule
UnTrac & -0.147±0.014 & -0.239±0.011 & 0.418±0.063 & 0.419±0.063 & 0.345±0.179 \\
UnTrac-Inv & -0.100±0.069 & -0.099±0.070 & -0.231±0.012 & 0.376±0.008 & 0.313±0.003 \\
\bottomrule
\toprule
Learning Rate & 5e-06 & 1e-05 & 5e-05 & 1e-04 & 5e-04 \\
\midrule
UnTrac & -0.127±0.302 & 0.312±0.311 & 0.419±0.063 & 0.377±0.040 & 0.329±0.015 \\
UnTrac-Inv & 0.100±0.084 & 0.197±0.067 & 0.376±0.008 & 0.137±0.019 & 0.027±0.015 \\
\bottomrule
\end{tabularx}
\caption{Hyperparameter sensitivity across different optimizers and learning rates. 
Each figure denotes the Pearson correlation coefficient between the estimated and ground-truth influences.
The mean and standard deviation across four runs are displayed.}
\label{tbl:hyperparameter}
\end{table*}

\subsection{Sensitivity to Optimizer \& Learning Rate}
\label{sec:hyperparameter}

Here, we discuss how the choice of optimizers and learning rate affects the performance of our methods.
Table \ref{tbl:hyperparameter} shows the performance of each method across different optimizers and learning rates, while fixing other hyperparameters.

\paragraph{Optimizer}

Table \ref{tbl:hyperparameter} (top) shows the performance with various optimizers.
We use SGD, SGD with momentum ($\mathrm{momentum}\!=\!\mathrm{dampening}\!=\!0.9$), RMSProp (\citealp{tieleman2012lecture}; $\alpha=0.99$), Adam, and Adafactor \cite{shazeer2018adafactor}.
UnTrac performs well when RMSProp, Adam, and Adafactor are used, indicating that a preconditioner plays an important role in unlearning.

A similar trend can be seen for UnTrac-Inv, though it performs worse when RMSProp is used.
We suspect this is because both Adam and Adafactor use a decaying average of gradients, rather than just the gradient for the current batch as RMSProp and SGD do. 
This makes training less stochastic and has a somewhat similar effect to using larger batches, which is beneficial for UnTrac-Inv.

\paragraph{Learning Rate}

Table \ref{tbl:hyperparameter} (bottom) presents the performance with several learning rates.
UnTrac performs well across various learning rates when using higher learning rates.
With lower learning rates, UnTrac does not converge and performs unstably. 

In contrast, UnTrac-Inv is somewhat sensitive to the choice of learning rate.
Following the discussion in Section \ref{sec:untrac_reverse}, UnTrac-Inv diverges from UnTrac as unlearning proceeds. 
As a higher learning rate boosts the divergence, increasing the learning rate does not necessarily lead to higher performance.
However, the estimated influence correlates positively with ground truth on any learning rate.

\section{Related Work}

Which training data affects the model’s prediction? 
The importance of this question has been rising along with the emergence of LLMs trained on massive text corpora \cite{akyurek-etal-2022-towards, longpre2023pretrainer, feng-etal-2023-pretraining, han-etal-2023-understanding, grosse2023studying}.
Hessian-based influence functions \cite{hampel1974influence, koh2017understanding} are pioneering work to address this question and are widely used in the context of natural language processing \cite{han-etal-2020-explaining, guo-etal-2021-fastif, yang-etal-2023-many}.
As computing inverse Hessian is quite expensive \cite{guo-etal-2021-fastif, schioppa2022scaling, grosse2023studying}, TracIn is another promising approach for training data attribution \cite{pruthi2020estimating}.
However, several studies empirically reported that these methods are unstable for deep neural networks \cite{basu2021influence, sogaard2021revisiting}.

While the studies above focus on a single training example, several studies consider the influences of multiple training examples \cite{koh2019accuracy, barshan2020relatif}.
\citet{basu2020second} argue that the first-order approximation behind influence functions is unsuitable, as removing multiple examples leads to a significant perturbation of model parameters.
\citet{schioppa2023theoretical} theoretically show that influence functions work well for a limited number of training steps, but are unlikely to be good approximators of leave-one-out influences.
Therefore, they used influence functions to detect influential training examples of mispredictions.
By flipping the label of the influential training examples and training a model on the flipped examples, the mispredictions can be corrected with a few finetuning steps.
Whereas they use influence functions to detect influential examples to be unlearned, we use unlearning to detect influential examples.

Sampling-based attribution methods, such as Shapley values \cite{ghorbani2019data}, empirical influences \cite{feldman2020neural}, and datamodels \cite{ilyas2022datamodels} are also promising approaches for tracing influential training datasets.
However, they commonly require thousands of pairs of training subsets and models trained on the subsets to estimate how the combination of training examples affects the model’s performance. 
Although TRAK \cite{park2023trak}, a recent sampling-based approach, requires significantly fewer computational resources than others, it still requires tens of pairs of training subsets and their corresponding trained models, which can be costly as well as leave-dataset-out. Indeed, TRAK was evaluated only on fine-tuning setups in its original paper. 
On the other hand, we study approaches that are scalable to large pretraining datasets.

Unlearning has been studied to remove undesired data from a trained model \cite{cao2015towards, golatkar2020eternal, sekhari2021remember, gupta2021adaptive} and has recently been used for LLMs \cite{jang-etal-2023-knowledge, wang-etal-2023-kga, yao2023large, maini2024tofu, zhang2024negative, jia2024soul}.

\section{Conclusion}

This paper explored unlearning approaches to trace the influence of training datasets.
Our approach, UnTrac, simply unlearns each training dataset and evaluates how the model's performance changes on a test dataset after unlearning.
UnTrac-Inv is a more scalable approach, which unlearns a test dataset and evaluates the unlearned model's performance on training datasets. 
Experimental results showed that our methods can trace the influence of training datasets with significantly higher accuracy than existing methods.
UnTrac works robustly if we use preconditioned optimizers with high learning rates and a sufficient number of training steps.
In contrast, UnTrac-Inv is relatively sensitive to the choice of hyperparameters, such as batch size, learning rate, and the number of training iterations.

As our methods require the same memory footprint as standard training, they can be applied to LLMs as long as we have enough memory space for training.
In future work, we hope that the effectiveness of our methods will be investigated with even more large models, and that our methods will be helpful in revealing the source of LLM's emergent abilities, such as chain-of-thought reasoning.

\section*{Limitations}

Although our methods are assessed across different model architectures, model sizes, and training setups, further empirical investigations are expected to verify their effectiveness.
We do not argue that our methods are immediately applicable in practical use.
For real-world applications, we need to validate our methods in various settings and tasks against leave-dataset-out, though leave-dataset-out requires considerable computational cost.


\section*{Acknowledgements}

MI is partially supported by JST CREST JPMJCR21D1, NEDO JPNP20006, and JSPS KAKENHI 23K16940, Japan.
IT is supported by the Dutch National Science Foundation (NWO Vici VI.C.212.053).

\bibliography{anthology1, custom}

\begin{thebibliography}{54}
\expandafter\ifx\csname natexlab\endcsname\relax\def\natexlab#1{#1}\fi

\bibitem[{Agarwal et~al.(2017)Agarwal, Bullins, and Hazan}]{agarwal2017second}
Naman Agarwal, Brian Bullins, and Elad Hazan. 2017.
\newblock Second-order stochastic optimization for machine learning in linear time.
\newblock \emph{Journal of Machine Learning Research}, 18(1):4148--4187.

\bibitem[{Akyurek et~al.(2022)Akyurek, Bolukbasi, Liu, Xiong, Tenney, Andreas, and Guu}]{akyurek-etal-2022-towards}
Ekin Akyurek, Tolga Bolukbasi, Frederick Liu, Binbin Xiong, Ian Tenney, Jacob Andreas, and Kelvin Guu. 2022.
\newblock \href {https://doi.org/10.18653/v1/2022.findings-emnlp.180} {Towards tracing knowledge in language models back to the training data}.
\newblock In \emph{Findings of the Association for Computational Linguistics: EMNLP 2022}, pages 2429--2446, Abu Dhabi, United Arab Emirates. Association for Computational Linguistics.

\bibitem[{Barshan et~al.(2020)Barshan, Brunet, and Dziugaite}]{barshan2020relatif}
Elnaz Barshan, Marc-Etienne Brunet, and Gintare~Karolina Dziugaite. 2020.
\newblock Relatif: Identifying explanatory training samples via relative influence.
\newblock In \emph{International Conference on Artificial Intelligence and Statistics}, pages 1899--1909. PMLR.

\bibitem[{Basu et~al.(2021)Basu, Pope, and Feizi}]{basu2021influence}
Samyadeep Basu, Phil Pope, and Soheil Feizi. 2021.
\newblock Influence functions in deep learning are fragile.
\newblock In \emph{International Conference on Learning Representations}.

\bibitem[{Basu et~al.(2020)Basu, You, and Feizi}]{basu2020second}
Samyadeep Basu, Xuchen You, and Soheil Feizi. 2020.
\newblock On second-order group influence functions for black-box predictions.
\newblock In \emph{Proceedings of the 37th International Conference on Machine Learning}, pages 715--724. PMLR.

\bibitem[{Cao and Yang(2015)}]{cao2015towards}
Yinzhi Cao and Junfeng Yang. 2015.
\newblock Towards making systems forget with machine unlearning.
\newblock In \emph{IEEE Symposium on Security and Privacy}, pages 463--480. IEEE.

\bibitem[{Chen and Yang(2023)}]{chen-yang-2023-unlearn}
Jiaao Chen and Diyi Yang. 2023.
\newblock \href {https://doi.org/10.18653/v1/2023.emnlp-main.738} {Unlearn what you want to forget: Efficient unlearning for {LLM}s}.
\newblock In \emph{Proceedings of the 2023 Conference on Empirical Methods in Natural Language Processing}, pages 12041--12052, Singapore. Association for Computational Linguistics.

\bibitem[{Cook and Weisberg(1982)}]{cook1982residuals}
R~Dennis Cook and Sanford Weisberg. 1982.
\newblock \emph{Residuals and Influence in Regression}.
\newblock New York: Chapman and Hall.

\bibitem[{Feldman and Zhang(2020)}]{feldman2020neural}
Vitaly Feldman and Chiyuan Zhang. 2020.
\newblock What neural networks memorize and why: Discovering the long tail via influence estimation.
\newblock \emph{Advances in Neural Information Processing Systems}, 33:2881--2891.

\bibitem[{Feng et~al.(2023)Feng, Park, Liu, and Tsvetkov}]{feng-etal-2023-pretraining}
Shangbin Feng, Chan~Young Park, Yuhan Liu, and Yulia Tsvetkov. 2023.
\newblock \href {https://doi.org/10.18653/v1/2023.acl-long.656} {From pretraining data to language models to downstream tasks: Tracking the trails of political biases leading to unfair {NLP} models}.
\newblock In \emph{Proceedings of the 61st Annual Meeting of the Association for Computational Linguistics (Volume 1: Long Papers)}, pages 11737--11762, Toronto, Canada. Association for Computational Linguistics.

\bibitem[{Gao et~al.(2020)Gao, Biderman, Black, Golding, Hoppe, Foster, Phang, He, Thite, Nabeshima et~al.}]{gao2020pile}
Leo Gao, Stella Biderman, Sid Black, Laurence Golding, Travis Hoppe, Charles Foster, Jason Phang, Horace He, Anish Thite, Noa Nabeshima, et~al. 2020.
\newblock The pile: An 800gb dataset of diverse text for language modeling.
\newblock \emph{arXiv preprint arXiv:2101.00027}.

\bibitem[{Ghorbani and Zou(2019)}]{ghorbani2019data}
Amirata Ghorbani and James Zou. 2019.
\newblock Data shapley: Equitable valuation of data for machine learning.
\newblock In \emph{International conference on machine learning}, pages 2242--2251. PMLR.

\bibitem[{Ginart et~al.(2019)Ginart, Guan, Valiant, and Zou}]{ginart2019making}
Antonio Ginart, Melody Guan, Gregory Valiant, and James~Y Zou. 2019.
\newblock Making ai forget you: Data deletion in machine learning.
\newblock \emph{Advances in neural information processing systems}, 32.

\bibitem[{Golatkar et~al.(2020)Golatkar, Achille, and Soatto}]{golatkar2020eternal}
Aditya Golatkar, Alessandro Achille, and Stefano Soatto. 2020.
\newblock Eternal sunshine of the spotless net: Selective forgetting in deep networks.
\newblock In \emph{Proceedings of the IEEE/CVF Conference on Computer Vision and Pattern Recognition}, pages 9304--9312.

\bibitem[{Grosse et~al.(2023)Grosse, Bae, Anil, Elhage, Tamkin, Tajdini, Steiner, Li, Durmus, Perez et~al.}]{grosse2023studying}
Roger Grosse, Juhan Bae, Cem Anil, Nelson Elhage, Alex Tamkin, Amirhossein Tajdini, Benoit Steiner, Dustin Li, Esin Durmus, Ethan Perez, et~al. 2023.
\newblock Studying large language model generalization with influence functions.
\newblock \emph{arXiv preprint arXiv:2308.03296}.

\bibitem[{Guo et~al.(2021)Guo, Rajani, Hase, Bansal, and Xiong}]{guo-etal-2021-fastif}
Han Guo, Nazneen Rajani, Peter Hase, Mohit Bansal, and Caiming Xiong. 2021.
\newblock \href {https://doi.org/10.18653/v1/2021.emnlp-main.808} {{F}ast{IF}: Scalable influence functions for efficient model interpretation and debugging}.
\newblock In \emph{Proceedings of the 2021 Conference on Empirical Methods in Natural Language Processing}, pages 10333--10350, Online and Punta Cana, Dominican Republic. Association for Computational Linguistics.

\bibitem[{Gupta et~al.(2021)Gupta, Jung, Neel, Roth, Sharifi-Malvajerdi, and Waites}]{gupta2021adaptive}
Varun Gupta, Christopher Jung, Seth Neel, Aaron Roth, Saeed Sharifi-Malvajerdi, and Chris Waites. 2021.
\newblock Adaptive machine unlearning.
\newblock \emph{Advances in Neural Information Processing Systems}, 34:16319--16330.

\bibitem[{Hampel(1974)}]{hampel1974influence}
Frank~R Hampel. 1974.
\newblock The influence curve and its role in robust estimation.
\newblock \emph{Journal of the American Statistical Association}, 69(346):383--393.

\bibitem[{Han et~al.(2023)Han, Simig, Mihaylov, Tsvetkov, Celikyilmaz, and Wang}]{han-etal-2023-understanding}
Xiaochuang Han, Daniel Simig, Todor Mihaylov, Yulia Tsvetkov, Asli Celikyilmaz, and Tianlu Wang. 2023.
\newblock \href {https://doi.org/10.18653/v1/2023.acl-long.708} {Understanding in-context learning via supportive pretraining data}.
\newblock In \emph{Proceedings of the 61st Annual Meeting of the Association for Computational Linguistics (Volume 1: Long Papers)}, pages 12660--12673, Toronto, Canada. Association for Computational Linguistics.

\bibitem[{Han and Tsvetkov(2021)}]{han-tsvetkov-2021-influence-tuning}
Xiaochuang Han and Yulia Tsvetkov. 2021.
\newblock \href {https://doi.org/10.18653/v1/2021.findings-emnlp.374} {Influence tuning: Demoting spurious correlations via instance attribution and instance-driven updates}.
\newblock In \emph{Findings of the Association for Computational Linguistics: EMNLP 2021}, pages 4398--4409, Punta Cana, Dominican Republic. Association for Computational Linguistics.

\bibitem[{Han et~al.(2020)Han, Wallace, and Tsvetkov}]{han-etal-2020-explaining}
Xiaochuang Han, Byron~C. Wallace, and Yulia Tsvetkov. 2020.
\newblock \href {https://doi.org/10.18653/v1/2020.acl-main.492} {Explaining black box predictions and unveiling data artifacts through influence functions}.
\newblock In \emph{Proceedings of the 58th Annual Meeting of the Association for Computational Linguistics}, pages 5553--5563, Online. Association for Computational Linguistics.

\bibitem[{Hartvigsen et~al.(2022)Hartvigsen, Gabriel, Palangi, Sap, Ray, and Kamar}]{hartvigsen-etal-2022-toxigen}
Thomas Hartvigsen, Saadia Gabriel, Hamid Palangi, Maarten Sap, Dipankar Ray, and Ece Kamar. 2022.
\newblock \href {https://doi.org/10.18653/v1/2022.acl-long.234} {{T}oxi{G}en: A large-scale machine-generated dataset for adversarial and implicit hate speech detection}.
\newblock In \emph{Proceedings of the 60th Annual Meeting of the Association for Computational Linguistics (Volume 1: Long Papers)}, pages 3309--3326, Dublin, Ireland. Association for Computational Linguistics.

\bibitem[{Ilyas et~al.(2022)Ilyas, Park, Engstrom, Leclerc, and Madry}]{ilyas2022datamodels}
Andrew Ilyas, Sung~Min Park, Logan Engstrom, Guillaume Leclerc, and Aleksander Madry. 2022.
\newblock Datamodels: Understanding predictions with data and data with predictions.
\newblock In \emph{Proceedings of the 39th International Conference on Machine Learning}, pages 9525--9587. PMLR.

\bibitem[{Jang et~al.(2023)Jang, Yoon, Yang, Cha, Lee, Logeswaran, and Seo}]{jang-etal-2023-knowledge}
Joel Jang, Dongkeun Yoon, Sohee Yang, Sungmin Cha, Moontae Lee, Lajanugen Logeswaran, and Minjoon Seo. 2023.
\newblock \href {https://doi.org/10.18653/v1/2023.acl-long.805} {Knowledge unlearning for mitigating privacy risks in language models}.
\newblock In \emph{Proceedings of the 61st Annual Meeting of the Association for Computational Linguistics (Volume 1: Long Papers)}, pages 14389--14408, Toronto, Canada. Association for Computational Linguistics.

\bibitem[{Jia et~al.(2024)Jia, Zhang, Zhang, Liu, Runwal, Diffenderfer, Kailkhura, and Liu}]{jia2024soul}
Jinghan Jia, Yihua Zhang, Yimeng Zhang, Jiancheng Liu, Bharat Runwal, James Diffenderfer, Bhavya Kailkhura, and Sijia Liu. 2024.
\newblock Soul: Unlocking the power of second-order optimization for llm unlearning.
\newblock \emph{arXiv preprint arXiv:2404.18239}.

\bibitem[{Kingma and Ba(2014)}]{kingma2014adam}
Diederik~P Kingma and Jimmy Ba. 2014.
\newblock Adam: A method for stochastic optimization.
\newblock \emph{arXiv preprint arXiv:1412.6980v9}.

\bibitem[{Koh and Liang(2017)}]{koh2017understanding}
Pang~Wei Koh and Percy Liang. 2017.
\newblock Understanding black-box predictions via influence functions.
\newblock In \emph{Proceedings of the 34th International Conference on Machine Learning}, pages 1885--1894. PMLR.

\bibitem[{Koh et~al.(2019)Koh, Ang, Teo, and Liang}]{koh2019accuracy}
Pang Wei~W Koh, Kai-Siang Ang, Hubert Teo, and Percy~S Liang. 2019.
\newblock On the accuracy of influence functions for measuring group effects.
\newblock \emph{Advances in Neural Information Processing Systems}, 32.

\bibitem[{Lester et~al.(2021)Lester, Al-Rfou, and Constant}]{lester-etal-2021-power}
Brian Lester, Rami Al-Rfou, and Noah Constant. 2021.
\newblock \href {https://doi.org/10.18653/v1/2021.emnlp-main.243} {The power of scale for parameter-efficient prompt tuning}.
\newblock In \emph{Proceedings of the 2021 Conference on Empirical Methods in Natural Language Processing}, pages 3045--3059, Online and Punta Cana, Dominican Republic. Association for Computational Linguistics.

\bibitem[{Lin et~al.(2022)Lin, Hilton, and Evans}]{lin-etal-2022-truthfulqa}
Stephanie Lin, Jacob Hilton, and Owain Evans. 2022.
\newblock \href {https://doi.org/10.18653/v1/2022.acl-long.229} {{T}ruthful{QA}: Measuring how models mimic human falsehoods}.
\newblock In \emph{Proceedings of the 60th Annual Meeting of the Association for Computational Linguistics (Volume 1: Long Papers)}, pages 3214--3252, Dublin, Ireland. Association for Computational Linguistics.

\bibitem[{Liu et~al.(2019)Liu, Ott, Goyal, Du, Joshi, Chen, Levy, Lewis, Zettlemoyer, and Stoyanov}]{liu2019roberta}
Yinhan Liu, Myle Ott, Naman Goyal, Jingfei Du, Mandar Joshi, Danqi Chen, Omer Levy, Mike Lewis, Luke Zettlemoyer, and Veselin Stoyanov. 2019.
\newblock Roberta: A robustly optimized bert pretraining approach.
\newblock \emph{arXiv preprint arXiv:1907.11692}.

\bibitem[{Longpre et~al.(2023)Longpre, Yauney, Reif, Lee, Roberts, Zoph, Zhou, Wei, Robinson, Mimno et~al.}]{longpre2023pretrainer}
Shayne Longpre, Gregory Yauney, Emily Reif, Katherine Lee, Adam Roberts, Barret Zoph, Denny Zhou, Jason Wei, Kevin Robinson, David Mimno, et~al. 2023.
\newblock A pretrainer's guide to training data: Measuring the effects of data age, domain coverage, quality, \& toxicity.
\newblock \emph{arXiv preprint arXiv:2305.13169}.

\bibitem[{Maini et~al.(2024)Maini, Feng, Schwarzschild, Lipton, and Kolter}]{maini2024tofu}
Pratyush Maini, Zhili Feng, Avi Schwarzschild, Zachary~C Lipton, and J~Zico Kolter. 2024.
\newblock Tofu: A task of fictitious unlearning for llms.
\newblock \emph{arXiv preprint arXiv:2401.06121}.

\bibitem[{Mehta et~al.(2022)Mehta, Pal, Singh, and Ravi}]{mehta2022deep}
Ronak Mehta, Sourav Pal, Vikas Singh, and Sathya~N Ravi. 2022.
\newblock Deep unlearning via randomized conditionally independent hessians.
\newblock In \emph{Proceedings of the IEEE/CVF Conference on Computer Vision and Pattern Recognition}, pages 10422--10431.

\bibitem[{Miller(1994)}]{miller-1994-wordnet}
George~A. Miller. 1994.
\newblock \href {https://aclanthology.org/H94-1111} {{W}ord{N}et: A lexical database for {E}nglish}.
\newblock In \emph{{H}uman {L}anguage {T}echnology: Proceedings of a Workshop held at {P}lainsboro, {N}ew {J}ersey, {M}arch 8-11, 1994}.

\bibitem[{Park et~al.(2023)Park, Georgiev, Ilyas, Leclerc, and M{\k{a}}dry}]{park2023trak}
Sung~Min Park, Kristian Georgiev, Andrew Ilyas, Guillaume Leclerc, and Aleksander M{\k{a}}dry. 2023.
\newblock Trak: attributing model behavior at scale.
\newblock In \emph{Proceedings of the 40th International Conference on Machine Learning}, pages 27074--27113. PMLR.

\bibitem[{Paszke et~al.(2019)Paszke, Gross, Massa, Lerer, Bradbury, Chanan, Killeen, Lin, Gimelshein, Antiga, Desmaison, Kopf, Yang, DeVito, Raison, Tejani, Chilamkurthy, Steiner, Fang, Bai, and Chintala}]{NEURIPS2019_9015}
Adam Paszke, Sam Gross, Francisco Massa, Adam Lerer, James Bradbury, Gregory Chanan, Trevor Killeen, Zeming Lin, Natalia Gimelshein, Luca Antiga, Alban Desmaison, Andreas Kopf, Edward Yang, Zachary DeVito, Martin Raison, Alykhan Tejani, Sasank Chilamkurthy, Benoit Steiner, Lu~Fang, Junjie Bai, and Soumith Chintala. 2019.
\newblock Pytorch: An imperative style, high-performance deep learning library.
\newblock In \emph{Advances in Neural Information Processing Systems}, pages 8024--8035. Curran Associates, Inc.

\bibitem[{Pruthi et~al.(2020)Pruthi, Liu, Kale, and Sundararajan}]{pruthi2020estimating}
Garima Pruthi, Frederick Liu, Satyen Kale, and Mukund Sundararajan. 2020.
\newblock Estimating training data influence by tracing gradient descent.
\newblock \emph{Advances in Neural Information Processing Systems}, 33:19920--19930.

\bibitem[{Raffel et~al.(2020)Raffel, Shazeer, Roberts, Lee, Narang, Matena, Zhou, Li, Liu et~al.}]{raffel2020exploring}
Colin Raffel, Noam Shazeer, Adam Roberts, Katherine Lee, Sharan Narang, Michael Matena, Yanqi Zhou, Wei Li, Peter~J Liu, et~al. 2020.
\newblock Exploring the limits of transfer learning with a unified text-to-text transformer.
\newblock \emph{Journal of Machine Learning Research}, 21(140):1--67.

\bibitem[{Schioppa et~al.(2023)Schioppa, Filippova, Titov, and Zablotskaia}]{schioppa2023theoretical}
Andrea Schioppa, Katja Filippova, Ivan Titov, and Polina Zablotskaia. 2023.
\newblock Theoretical and practical perspectives on what influence functions do.
\newblock \emph{Advances in Neural Information Processing Systems}.

\bibitem[{Schioppa et~al.(2022)Schioppa, Zablotskaia, Vilar, and Sokolov}]{schioppa2022scaling}
Andrea Schioppa, Polina Zablotskaia, David Vilar, and Artem Sokolov. 2022.
\newblock Scaling up influence functions.
\newblock In \emph{Proceedings of the AAAI Conference on Artificial Intelligence}, pages 8179--8186.

\bibitem[{Sekhari et~al.(2021)Sekhari, Acharya, Kamath, and Suresh}]{sekhari2021remember}
Ayush Sekhari, Jayadev Acharya, Gautam Kamath, and Ananda~Theertha Suresh. 2021.
\newblock Remember what you want to forget: Algorithms for machine unlearning.
\newblock \emph{Advances in Neural Information Processing Systems}, 34:18075--18086.

\bibitem[{Shazeer and Stern(2018)}]{shazeer2018adafactor}
Noam Shazeer and Mitchell Stern. 2018.
\newblock Adafactor: Adaptive learning rates with sublinear memory cost.
\newblock In \emph{Proceedings of the 35th International Conference on Machine Learning}, pages 4596--4604. PMLR.

\bibitem[{S{\o}gaard(2021)}]{sogaard2021revisiting}
Anders S{\o}gaard. 2021.
\newblock Revisiting methods for finding influential examples.
\newblock \emph{arXiv preprint arXiv:2111.04683}.

\bibitem[{Tieleman et~al.(2012)Tieleman, Hinton et~al.}]{tieleman2012lecture}
Tijmen Tieleman, Geoffrey Hinton, et~al. 2012.
\newblock Lecture 6.5-rmsprop: Divide the gradient by a running average of its recent magnitude.
\newblock \emph{COURSERA: Neural networks for machine learning}, 4(2):26--31.

\bibitem[{Trinh and Le(2018)}]{trinh2018simple}
Trieu~H Trinh and Quoc~V Le. 2018.
\newblock A simple method for commonsense reasoning.
\newblock \emph{arXiv preprint arXiv:1806.02847}.

\bibitem[{Wang et~al.(2023)Wang, Chen, Yuan, Zeng, Wong, and Yin}]{wang-etal-2023-kga}
Lingzhi Wang, Tong Chen, Wei Yuan, Xingshan Zeng, Kam-Fai Wong, and Hongzhi Yin. 2023.
\newblock \href {https://doi.org/10.18653/v1/2023.acl-long.740} {{KGA}: A general machine unlearning framework based on knowledge gap alignment}.
\newblock In \emph{Proceedings of the 61st Annual Meeting of the Association for Computational Linguistics (Volume 1: Long Papers)}, pages 13264--13276, Toronto, Canada. Association for Computational Linguistics.

\bibitem[{Wolf et~al.(2020)Wolf, Debut, Sanh, Chaumond, Delangue, Moi, Cistac, Rault, Louf, Funtowicz, Davison, Shleifer, von Platen, Ma, Jernite, Plu, Xu, Le~Scao, Gugger, Drame, Lhoest, and Rush}]{wolf-etal-2020-transformers}
Thomas Wolf, Lysandre Debut, Victor Sanh, Julien Chaumond, Clement Delangue, Anthony Moi, Pierric Cistac, Tim Rault, Remi Louf, Morgan Funtowicz, Joe Davison, Sam Shleifer, Patrick von Platen, Clara Ma, Yacine Jernite, Julien Plu, Canwen Xu, Teven Le~Scao, Sylvain Gugger, Mariama Drame, Quentin Lhoest, and Alexander Rush. 2020.
\newblock \href {https://doi.org/10.18653/v1/2020.emnlp-demos.6} {Transformers: State-of-the-art natural language processing}.
\newblock In \emph{Proceedings of the 2020 Conference on Empirical Methods in Natural Language Processing: System Demonstrations}, pages 38--45, Online. Association for Computational Linguistics.

\bibitem[{Yang et~al.(2023)Yang, Jain, and Wallace}]{yang-etal-2023-many}
Jinghan Yang, Sarthak Jain, and Byron~C. Wallace. 2023.
\newblock \href {https://doi.org/10.18653/v1/2023.eacl-main.188} {How many and which training points would need to be removed to flip this prediction?}
\newblock In \emph{Proceedings of the 17th Conference of the European Chapter of the Association for Computational Linguistics}, pages 2571--2584, Dubrovnik, Croatia. Association for Computational Linguistics.

\bibitem[{Yao et~al.(2023)Yao, Xu, and Liu}]{yao2023large}
Yuanshun Yao, Xiaojun Xu, and Yang Liu. 2023.
\newblock Large language model unlearning.
\newblock \emph{arXiv preprint arXiv:2310.10683}.

\bibitem[{Zhang et~al.(2024)Zhang, Lin, Bai, and Mei}]{zhang2024negative}
Ruiqi Zhang, Licong Lin, Yu~Bai, and Song Mei. 2024.
\newblock Negative preference optimization: From catastrophic collapse to effective unlearning.
\newblock \emph{arXiv preprint arXiv:2404.05868}.

\bibitem[{Zhang et~al.(2022)Zhang, Roller, Goyal, Artetxe, Chen, Chen, Dewan, Diab, Li, Lin et~al.}]{zhang2022opt}
Susan Zhang, Stephen Roller, Naman Goyal, Mikel Artetxe, Moya Chen, Shuohui Chen, Christopher Dewan, Mona Diab, Xian Li, Xi~Victoria Lin, et~al. 2022.
\newblock Opt: Open pre-trained transformer language models.
\newblock \emph{arXiv preprint arXiv:2205.01068}.

\bibitem[{Zhao et~al.(2018)Zhao, Wang, Yatskar, Ordonez, and Chang}]{zhao-etal-2018-gender}
Jieyu Zhao, Tianlu Wang, Mark Yatskar, Vicente Ordonez, and Kai-Wei Chang. 2018.
\newblock \href {https://doi.org/10.18653/v1/N18-2003} {Gender bias in coreference resolution: Evaluation and debiasing methods}.
\newblock In \emph{Proceedings of the 2018 Conference of the North {A}merican Chapter of the Association for Computational Linguistics: Human Language Technologies, Volume 2 (Short Papers)}, pages 15--20, New Orleans, Louisiana. Association for Computational Linguistics.

\bibitem[{Zhu et~al.(2015)Zhu, Kiros, Zemel, Salakhutdinov, Urtasun, Torralba, and Fidler}]{zhu2015aligning}
Yukun Zhu, Ryan Kiros, Rich Zemel, Ruslan Salakhutdinov, Raquel Urtasun, Antonio Torralba, and Sanja Fidler. 2015.
\newblock Aligning books and movies: Towards story-like visual explanations by watching movies and reading books.
\newblock In \emph{Proceedings of the IEEE International Conference on Computer Vision}, pages 19--27.

\end{thebibliography}

\clearpage
\appendix

\begin{figure*}[t!]
\centering
\includegraphics[width=\linewidth]{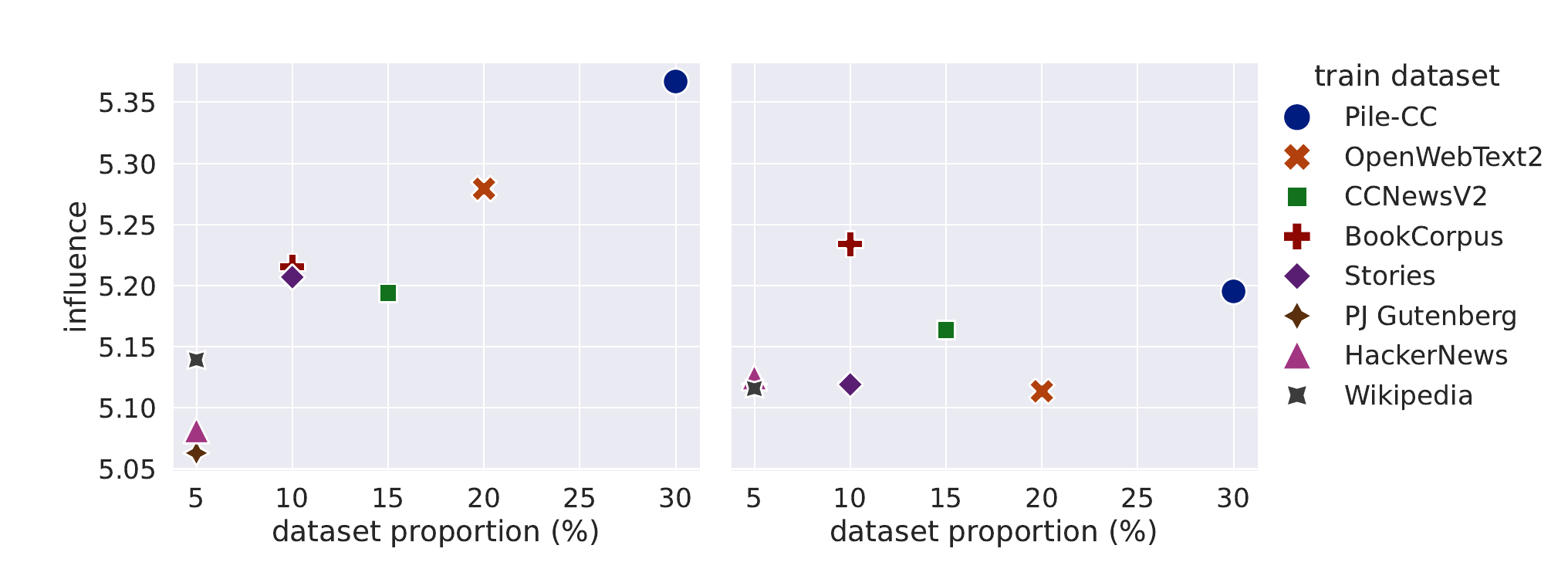}
\caption{Ground-truth influence computed by leave-dataset-out. Left: each counterfactual model is trained on all the examples except for a training dataset. Right: each counterfactual model is trained for the same number of steps.}
\label{fig:loo}
\end{figure*}

\begin{table*}[t!]
\small
\centering
\begin{tabular}{lrrrrrrrr}
\toprule
&\multicolumn{3}{c}{Equal Training Dataset Size}& &\multicolumn{3}{c}{Different Training Dataset Size}\\ 
\cmidrule{2-4} \cmidrule{6-8}
Test Dataset & ToxiGen & WinoBias & TruthfulQA & & ToxiGen & WinoBias & TruthfulQA \\
\midrule
GradDot & -0.017±0.010 & 0.390±0.020 & 0.122±0.014 &  & -0.116±0.008 & 0.445±0.051 & -0.483±0.010 \\
GradCos & 0.104±0.013 & 0.554±0.015 & 0.446±0.028 &  & -0.124±0.004 & 0.478±0.050 & -0.331±0.013 \\
HIF (Arnoldi) & -0.194±0.026 & 0.509±0.018 & 0.218±0.009 &  & -0.231±0.007 & 0.513±0.009 & -0.243±0.015 \\
HIF (LISSA) & -0.002±0.296 & 0.374±0.149 & -0.191±0.199 &  & 0.009±0.024 & -0.110±0.130 & -0.127±0.024 \\
TracIn & 0.337±0.005 & 0.039±0.030 & 0.574±0.016 &  & -0.037±0.009 & 0.201±0.003 & 0.020±0.043 \\
\midrule
UnTrac & 0.421±0.082 & 0.814±0.060 & 0.374±0.155 &  & 0.440±0.046 & 0.265±0.149 & 0.144±0.099 \\
UnTrac-Inv & 0.118±0.018 & 0.854±0.010 & 0.571±0.025 &  & 0.284±0.058 & 0.249±0.148 & 0.265±0.014 \\
\bottomrule
\end{tabular}
\caption{Spearman correlation coefficient between the influence estimated by each method and the ground-truth influence computed by leave-dataset-out.
Each figure denotes the mean and standard deviation across four runs.
For each run, we use different examples randomly sampled from the training dataset to compute its influence.}
\label{tbl:spearmanr}
\end{table*}

\section{Appendix}
\label{sec:appendix}

\subsection{How to Compute Leave-Dataset-Out}
\label{apx:loo}

Figure \ref{fig:loo} shows the ground-truth influence assessed by leave-dataset-out where the pertaining dataset sizes are varied and ToxiGen is used as a test dataset (see Section \ref{sec:pretraining}). 
Leave-dataset-out assumes a counterfactual model that is trained on the mixture of all training datasets $D$ except for a dataset $\mathcal{Z}$. 
In the left figure, the counterfactual model is trained on all the examples except for the training dataset $\mathcal{Z}$: $\bm{\theta}_{-\mathcal{Z}} \!=\! \mathrm{arg} \min_{\bm{\theta}} \sum_{z \in \mathcal{D} \setminus \mathcal{Z}} L(z, \bm{\theta})$. 
The influence of large datasets is higher because the size of $\mathcal{D} \setminus \mathcal{Z}$ becomes smaller, and the performance of model $\bm{\theta}_{-\mathcal{Z}}$ largely depends on the dataset sizes. 
In the right figure, each counterfactual model is trained for the same number of training steps $T$: $\bm{\theta}_{-\mathcal{Z}} = \mathrm{arg} \min_{\bm{\theta}} \sum_{t=1}^{T} L(z_t, \bm{\theta})$ where $z_t \sim \mathcal{D} \setminus \mathcal{Z}$.
This setup does not overestimate the influence of large datasets.
In a practical scenario, we want to figure out which training datasets should be used under a fixed computational resource.
Thus, the influence of each training dataset should be compared under the same number of training steps.

\subsection{Implementation Details}
\label{sec:implementation}

Our code is implemented with Python v3.8.13, PyTorch v1.12.0 \cite{NEURIPS2019_9015}, and Transformers v4.18.0 \cite{wolf-etal-2020-transformers}.
In Section \ref{sec:instructiontuning}, we used the synthetic datasets which are originally created in this study.
The numbers used in the output text range from 1 to 256, while the characters and their part of speech are collected from WordNet \cite[][WordNet 3.0 license]{miller-1994-wordnet}.
Our study was conducted under the licenses and terms of the scientific artifacts.

Our experiments were conducted with a single NVIDIA A100 (80GB) for each run in Section \ref{sec:instructiontuning} and a single Tesla V100 (16GB) for each run in Section \ref{sec:pretraining}.
Unlearning each training dataset (UnTrac) takes approximately two minutes in Section \ref{sec:instructiontuning} and four hours in Section \ref{sec:pretraining}.
Unlearning each test dataset (UnTrac-Inv) takes approximately one minute in Section \ref{sec:instructiontuning} and 30 minutes in Section \ref{sec:pretraining}, not including the time required for evaluation.

\subsection{Evaluation by Spearman Correlation}
\label{apx:spearmanr}

Table \ref{tbl:spearmanr} shows the Spearman's rank correlation between the estimated influence and the ground-truth influence computed by leave-dataset-out in Section \ref{sec:pretraining}.
The tendency is similar to the result shown in Table \ref{tbl:toxic_result}, which uses the Pearson correlation as an evaluation metric.
Across all datasets and settings, the estimated influence by UnTrac and UnTrac-Inv correlates well with the ground truth.

\end{document}